%% file: main.tex
\def\year{2021}\relax
\documentclass[letterpaper]{article} % DO NOT CHANGE THIS
\usepackage{aaai21}  % DO NOT CHANGE THIS
\usepackage{times}  % DO NOT CHANGE THIS
\usepackage{helvet} % DO NOT CHANGE THIS
\usepackage{courier}  % DO NOT CHANGE THIS
\usepackage[hyphens]{url}  % DO NOT CHANGE THIS
\usepackage{graphicx} % DO NOT CHANGE THIS
\usepackage{natbib}  % DO NOT CHANGE THIS AND DO NOT ADD ANY OPTIONS TO IT
\usepackage{caption} % DO NOT CHANGE THIS AND DO NOT ADD ANY OPTIONS TO IT
\graphicspath{./fig}
\usepackage{svg}
\usepackage[ruled]{algorithm2e}
\usepackage{amsfonts}
\usepackage{amsmath}

\usepackage{hyperref}
\usepackage{algorithmic}
\usepackage{todonotes}
\newcommand{\notefh}[1]{{\todo[inline]{FH: #1}}}
\renewcommand{\notefh}[1]{}

\title{Learning Synthetic Environments for Reinforcement Learning with Evolution Strategies}
\author{
    Fabio Ferreira\thanks{denotes equal contribution}\textsuperscript{\rm 1}, Thomas Nierhoff\footnotemark[1]\textsuperscript{\rm 1}, Frank Hutter\textsuperscript{\rm 1, \rm 2}
    \\
}
\affiliations{
    \textsuperscript{\rm 1} University of Freiburg
    \textsuperscript{\rm 2} Bosch Center for Artificial Intelligence
    \href{mailto:ferreira@cs.uni-freiburg.de}{\{ferreira, nierhoff, fh\}@cs.uni-freiburg.de}

}
\begin{document}

\maketitle

\begin{abstract}
This work explores learning agent-agnostic \emph{synthetic environments} (SEs) for Reinforcement Learning. SEs act as a proxy for target environments and allow agents to be trained more efficiently than when directly trained on the target environment. We formulate this as a bi-level optimization problem and represent an SE as a neural network. By using Natural Evolution Strategies and a population of SE parameter vectors, we train agents in the inner loop on evolving SEs while in the outer loop we use the performance on the target task as a score for meta-updating the SE population. We show empirically that our method is capable of learning SEs for two discrete-action-space tasks (CartPole-v0 and Acrobot-v1) that allow us to train agents more robustly and with up to 60\% fewer steps. 
\notefh{Actually, do you know what the compute requirements are? Each step is probably also faster with SEs, isn't it? That would, e.g., be important for applications in HPO/NAS, etc.}
Not only do we show in experiments with 4000 evaluations that the SEs are robust against hyperparameter changes such as the learning rate, batch sizes and network sizes, we also show that SEs trained with DDQN agents transfer in limited ways to a discrete-action-space version of TD3 and very well to Dueling DDQN.
\end{abstract}

\input{1_intro}

\input{2_method}
\input{3_experiments}
\input{4_conclusion}
\input{5_acknowledgements}

\footnotesize
\bibliography{bib/strings,bib/lib,bib/shortproc,bib/local}
%\bibliographystyle{aaa21}

%\clearpage
\input{6_appendix}
\end{document}

%% file: 1_intro.tex
\section{Introduction}
% What is the problem?
In this paper we consider the intriguing idea of learning a proxy data generating process for Reinforcement Learning (RL) that allows one to train learners more effectively and efficiently on a task, that is, to achieve similar or higher performance more quickly compared to when trained directly on the original data generating process. 
%Todo: references
The relevant literature is multifaceted with works in core-sets \citep{sener-iclr18a}, World Models \citep{ha-arxiv18a}, POET \citep{wang-arxiv19a}, Generative Teaching Networks \citep{such-icml20a}, Generative Playing Networks \citep{bontrager-arxiv20a}, and Reward Shaping \citep{zheng-neurips18a} which all constitute contributions addressing this idea.
% Why is it interesting?
Learning proxy models is a very promising direction, because their higher training and evaluation efficiency allows for new applications in fields such as AutoML~\cite{hutter-book19a-nonote}. Moreover, they can serve as a tool for algorithm and dataset design since the proxy can yield insights into the importance of passing states carrying large signal or by identifying information and underrepresented dataset classes required for efficient learning \citep{such-icml20a}. 

In this work, we focus on learning a data generating process for RL. More precisely, we investigate the question of whether we can learn a Markov Decision Process (MDP), which we refer to as a \emph{synthetic environment} (SE), that is capable of producing synthetic data to allow for effective and efficient teaching of a target task to an agent (learner) through an informed representation of the target environment. We report results on the (continuous-state and discrete-action-space) CartPole-v0 and Acrobot-v1 target tasks from OpenAI Gym ~\citep{brockman-arxiv16a} which show that our SEs can train different types of agents to perform well on the target tasks, and also that these agents can be trained more efficiently.

We approach this environment generation problem by posing it as a bi-level optimization problem. The inner loop trains the agent on an SE and, since we employ an agent-agnostic method, we can interchangeably adopt standard RL algorithms at will, for example, ones based on policy gradient \citep{sutton-neurips20a} or Q-learning \citep{q-learning}. In the outer loop, we assess the agent's performance by evaluating it on the real environment (target task). The collected reward is used as a score to update the SE parameters used in the inner loop. Here, we use a population of SE parameters and update them with Evolution Strategies~\citep{es1}. We employ the same learning strategy as in \citep{salimans-arxiv17a} which belongs to the family of Natural Evolution Strategies~\citep{wierstra-cec08a} but instead of optimizing over the agent policy parameter space we optimize over the SE parameter space.

We drew inspiration from Generative Teaching Networks ~\citep{such-icml20a}. While we similarly use a bi-level optimization scheme to learn a data generator, our approach is different in central aspects. Particularly, we do not use noise vectors as input to our SEs and view the posed question directly from the perspective of RL instead of Supervised Learning. Also, we use ES to avoid the need for explicitly computing second-order meta-gradients. While ES has its drawbacks, this is beneficial since explicitly computing second-order meta-gradients can be expensive and unstable \citep{metz-icml19a}, particularly in the RL setting where the length of the inner loop can be variant and high. ES can further easily be parallelized and enables our method to be agent-agnostic. Our contributions are as follows: We
\begin{itemize}
    \item show that learning of synthetic environments as a bi-level optimization problem with NES constitutes a feasible method that is capable of learning SEs for two discrete-action-space Gym tasks, CartPole-v0 and Acrobot-v1.
    \item show that SEs trained with DDQN agents are able to transfer to other agents, that is, very well to Dueling DDQN and in limited ways to TD3 (which we adapted to deal with discrete action spaces).
    \item provide empirical evidence that SEs, once generated, are efficient and robust in training agents, requiring up to 60\% fewer training steps while varying hyperparameters such as the learning rate, batch size and neural network size. Our code and trained SEs are made available publicly\footnote{\href{https://github.com/automl/learning_environments}{\url{https://github.com/automl/learning\_environments}}}.
    \item shed some light on what the agents learn from the synthetic environments in a small qualitative study.
\end{itemize}

%% file: 2_method.tex
\section{Method}

\subsection{Problem Statement}
We consider a Markov Decision Process represented by a 4-tuple $(\mathcal{S},\mathcal{A},\mathcal{P},\mathcal{R})$ with $\mathcal{S}$ as the set of states, $\mathcal{A}$ as the set of actions, $\mathcal{P}: \mathcal{S} \times \mathcal{A} \rightarrow \mathcal{S}$ as the transition probabilities between states if a specific action is executed in that state and $\mathcal{R}$ as the immediate rewards. The MDPs we will consider in this work are either  human-designed environments $\mathcal{E}_{real}$ (such as Gym environments) or learned synthetic environments $\mathcal{E}_{syn, \psi}$, referred to as \emph{SE}, represented by a neural network with the parameters $\psi$. Interfacing with the environments is in both cases almost identical: given an input $a \in \mathcal{A}$, the environment outputs a next state $s' \in \mathcal{S}$ and a reward $r_a(s,s')\in \mathcal{R}$. In the case of $\mathcal{E}_{syn, \psi}$, we additionally input the current state $s\in \mathcal{S}$ since we model it to be stateless.
%Now, let $F$ be an objective function parameterized by $\theta$. 
The central objective of an RL agent when interacting on an MDP $\mathcal{E}_{real}$ is to find an optimal policy $\pi_{\theta}$ parameterized by $\theta$ that maximizes the expected reward $F(\theta; \mathcal{E}_{real})$. In RL, there exist many different methods to optimize this objective, for example policy gradient \citep{sutton-neurips20a} or Q-Learning \citep{q-learning}.
We now consider the following bi-level optimization problem: find the parameters $\psi^{*}$, such that the policy $\pi_\theta$ found by an agent parameterized by $\theta$ that trains on $\mathcal{E}_{syn, \psi^{*}}$ will achieve the highest reward on a target environment $\mathcal{E}_{real}$.
%after being trained on $\mathcal{E}_{syn, \psi^{*}}$. 
Formally that is:
\begin{equation}
\begin{aligned}
& & \psi^{*} = \underset{\psi}{\text{arg max}}
& & & F(\theta^*(\psi); \mathcal{E}_{real}) \\
\text{s.t.} & & \theta^{*}(\psi) = \underset{\theta}{\text{arg max}}
& & & F(\theta; \mathcal{E}_{syn, \psi}). & &
\end{aligned}
\end{equation}
We can use standard RL algorithms for optimizing the agents on the SE in the inner loop. Although gradient-based optimization methods can be applied in the outer loop, we chose Natural Evolution Strategies (NES) over such methods to allow the optimization to be independent of the choice of the agent in the inner loop and to avoid computing potentially expensive, unstable, and agent-specific meta-gradients. Additional advantages of ES are that it is better suited for long episodes (which often occur in RL), sparse or delayed rewards \citep{salimans-arxiv17a}, and parallelization.

\subsection{Algorithm}
Based on the formulated problem statement, let us now explain our method. The overall NES scheme is adopted from \citet{salimans-arxiv17a} and depicted in Algorithm \ref{alg:es}. We instantiate the search distribution similarly as an isotropic multivariate Gaussian with mean 0 and a covariance $\sigma^2I$ yielding the score function estimator $\frac{1}{\sigma} \mathbb{E}_{\epsilon \sim N(0,I)}\{F(\psi + \sigma \epsilon)\epsilon\}$. The main difference to \citet{salimans-arxiv17a} is that, while they maintain a population over perturbed agent parameter vectors, our population consists of perturbed SE parameter vectors. In contrast to their approach, our NES approach also involves two optimizations, namely that of the agent and the SE parameters instead of only the agent parameters. Our algorithm first stochastically perturbs each population member according to the search distribution resulting in $\psi_i$. Then, a new randomly initialized agent is trained in ~\emph{TrainAgent} on the SE parameterized by $\psi_i$ for $n_e$ episodes. The trained agent with fixed parameters is then evaluated on the real environment in \emph{EvaluateAgent}, yielding the average cumulative reward across 10 test episodes which we use as a score $F_{\psi, i}$ in the above score function estimator. Finally, we update $\psi$ in \emph{UpdateSE} with a stochastic gradient estimate based on all member scores via a weighted sum $\psi \leftarrow \psi + \alpha \frac{1}{n_p \sigma} \sum^{n_p}_i F_i \epsilon_i$. We repeat this process $n_o$ times but perform manual early-stopping when a resulting SE is capable of training agents that consistently solve the target task. Finally, we use a parallel version of the algorithm, using one worker for each member of the population at the same time.

\setlength{\textfloatsep}{10pt}% Remove \textfloatsep
\begin{algorithm}[htb]
    Input: initial synthetic environment parameters $\psi$, real environment $\mathcal{E}_{real}$, $\epsilon_i \sim \mathcal{N}(0,\sigma^2 I)$, number of episodes $n_e$, population size $n_p$ \;
    \Repeat{$n_o$ steps}{ 
        \For{each member of the pop. $i = 1,2,\ldots,n_p$}{
            $\psi_i = \psi + \epsilon_i$ \; 
            \For{$n = 1,2,\ldots,n_{e}$}{ 
                $\theta_{i,n}$ $=$ TrainAgent($\theta_{i,n-1}, \mathcal{E}_{syn, \psi_i}$) \;
            }
            $F_{\psi,i}$ $=$ EvaluateAgent($\theta_{i,n_{e}}$, $\mathcal{E}_{real}$) \;
        }
        $\psi$ $\leftarrow$ UpdateSE($\psi$, $\big\{\epsilon_i\big\}_i$, $\big\{F_{\psi,i}\big\}_i$) \;
    }
\caption{\FuncSty{NES for Learning SEs} \label{alg:es}}
\end{algorithm}
% \vspace{-0.5cm}
\subsection{Heuristics for Agent Training and Evaluation}
Determining the number of required training episodes $n_e$ on an SE is challenging as the rewards of the SE may not provide information about the current agent's performance on the real environment. Thus, we use a heuristic to early-stop training once the agent's training performance on the \emph{SE} converged. Let us refer to the cumulative reward of the $k$-th training episode as $C_k$. The two values $\bar{C}_{d}$ and $\bar{C}_{2d}$ maintain a non-overlapping moving average of the cumulative rewards over the last $d$ and $2d$ respective episodes up to episode $k$. Now, if $\frac{|\bar{C}_d - \bar{C}_{2d}|}{|\bar{C}_{2d}|} \leq C_{diff}$ the training is stopped. In all experiments we choose $d=10$ and $C_{diff}=0.01$. Training agents on \emph{real environments} is stopped when the average cumulative reward across the last $d$ test episodes exceeds the solved reward threshold. In case no heuristic is triggered, we train for 1000 episodes at most on both env. types. Independent of which of the environments ($\mathcal{E}_{real}$ or $\mathcal{E}_{syn}$) we train an agent on, the process to assess the actual agent performance is equivalent: we do this by running the agent on 10 test episodes from $\mathcal{E}_{real}$ for a fixed number of task-specific steps (i.e. 200 on CartPole and 500 on Acrobot) and use the cumulative rewards for each episode as a performance proxy and evaluation data points for visualization.

\subsection{Agents, Hyperparameters, and Sampling}
% \vspace{-0.4cm}
While our agent-agnostic method in principle allows to train arbitrary RL agents, in this work we always use DDQN~\citep{hasselt-aaai16} for the training of our agents in Algorithm \ref{alg:es}. Instead, we study the robustness of SEs to varying agent hyperparameters and the transferability of the SEs to train other agents. For studying transferability, we use Dueling DDQN~\citep{wang-icml16a} and TD3~\citep{td3}. The latter is chosen because it does not solely rely on (deep) Q-Learning as it constitutes an algorithm of the actor-critic and policy gradient family. However, since TD3 is naturally designed to deal with continuous action spaces but both our tasks employ a discrete action space, we equip the actor with a Gumble-Softmax distribution~\citep{jang-iclr17a} with a learned temperature that enables us to operate on discrete actions while maintaining differentiability. Due to known sensitivity to hyperparameters (HPs), we apply a hyperparameter optimization for the execution of Algorithm \ref{alg:es}. In addition to the inner and outer loop of our algorithm, we use another outer loop to optimize some of the agent and NES HPs with BOHB~\citep{falkner-icml18a} to identify stable HPs. The optimized HPs are reported in Table \ref{tab:exp_synth_best_performance_opt} in the appendix. We did not optimize some of the HPs that would negatively affect runtime (e.g. population size, number of train and test episode, see Table \ref{tab:exp_synth_best_performance_fixed} in the appendix). Resulting from this optimization, we use mirrored sampling~\citep{brockhoff-ppsn10a} similar to \citet{salimans-arxiv17a} and a score transformation applied to $F$ that considers only members above the average of the population scores and which normalizes these to the range $[0, 1]$. In many of our experiments we draw a comparison between varying HPs (denoted as ``HP: varying'') and keeping HPs fixed (denoted as ``HP: fixed''). For the latter we use the default HPs specified in Table \ref{tab:default_hps} in the appendix. In the case of varying HPs, we focus on a subset of the default HPs given in Table \ref{tab:vary_hp} and randomly sample from the specified ranges.

\begin{table}[t]
\centering
\begin{tabular}{ |c|c|c| }  
 \hline
 agent hyperparameter & value range & log scale \\ 
 \hline
 learning rate & $10^{-3}/3$ - $10^{-3}*3$ & True \\ 
 batch size & $42$ - $384$ & True \\ 
 hidden size & $42$ - $384$ & True \\ 
 hidden layer & $1$ - $3$ & False \\
 \hline
\end{tabular} \par
\caption{Ranges used for sampling random agent HP config. for variation during NES runs and when training on SEs.}
\label{tab:vary_hp}
\end{table}

%% file: 3_experiments.tex
\section{Experiments}
After identifying stable DDQN and NES hyperparameters (HPs), we ran Algorithm \ref{alg:es} in parallel with 16 workers for 200 NES outer loop iterations. Each worker had one AMD EPYC 7502 CPU core at its disposal, resulting in an overall runtime of ~6-7h on Acrobot and ~5-6h on CartPole for 200 NES outer loop iterations. For reference, we note that \citet{salimans-arxiv17a} used 80 workers each having access to 18 cores for solving the Humanoid task.

\begin{figure}
\begin{center}
\centering
\includegraphics[width=1.0\linewidth]{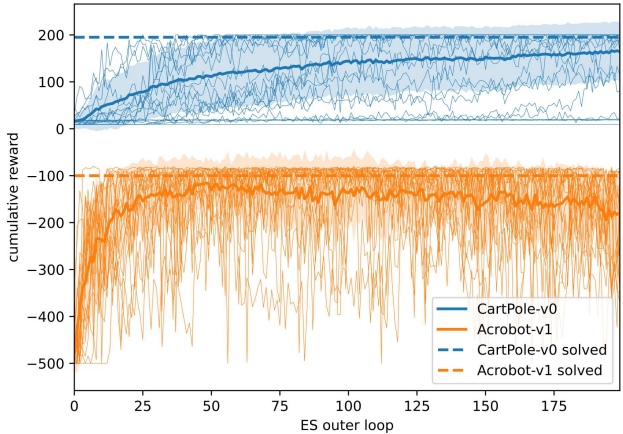} 
\end{center}
\vspace*{-0.2cm}
\caption{Results from 40 different NES runs with 16 workers each (using random seeds) show that our method is able to identify SEs that allow agents to solve the target tasks. Each thin line corresponds to the average of 16 worker evaluation scores returned by \emph{EvaluateAgent} in our algorithm as a function of the NES outer loop iterations.}
\label{fig:cartpole_acrobot_success}
\end{figure}

Let us now consider Figure \ref{fig:cartpole_acrobot_success}. It becomes evident that the proposed method with the given resources allows to identify SEs that are able to teach agents to solve CartPole and Acrobot tasks respectively. Each thin line in the plot corresponds to the average of 16 worker evaluation scores given by \emph{EvaluateAgent} in Algorithm \ref{alg:es} as a function of the NES outer loop iteration. We repeat this for 40 separate NES optimization runs and visualize the average across the thin lines as a thick line for each task. We note that we only show a random subset of the 40 thin lines for better visualization and randomly sample the seeds at all times in this work. We believe that the stochasticity introduced by this may lead to the variance visible in the plot when searching for good SEs. Both the stochasticity of natural gradients and the sensitivity of RL agents to seeds and parameter initializations may additionally contribute to this effect. Notice, it is often sufficient to run approximately 50 NES outer loops in order to find SEs that solve a task, as can be seen in Figure \ref{fig:cartpole_acrobot_success}. Besides early-stopping, other regularization techniques (e.g. regularizing the SE parameters) can be applied to address overfitting which we likely observe in the advanced training of the Acrobot task.
% DDQN HPs not varied during training of SE in Plot 1
% runtimes:
%../results/GTNC_evaluate_cartpole_2020-12-04-12
% mean [s]: 19600.585989597366
% std [s]: 6159.4935115799435
% ../results/GTNC_evaluate_acrobot_2020-11-28-16
% mean [s]: 22096.616291999817
% std [s]: 5844.505082061685

\subsection{Evaluation Performance of SE-trained Agents}
To answer whether the proposed method can learn SEs that are capable training agents effectively and efficiently, we conducted the following experiment. 

First, we generated two sets each consisting of 40 SEs from individual NES runs all capable of solving the CartPole task (i.e. a cumulative reward of $\geq 195$). For one of the sets we generated 40 SEs where we varied the HPs by sampling configurations of a subset of the DDQN HPs according to Table \ref{tab:vary_hp} before running the inner loop. For the other set of 40 SEs we did not vary the HPs and used our default DDQN HPs (see Table \ref{tab:default_hps} in the appendix). 

Second, on each SE of both of the sets we trained 10 DDQN agents with again varying HPs (again in the ranges specified in Table \ref{tab:vary_hp}). Then, after training, we evaluated each agent on the target task across 10 test episodes, resulting in overall 4000 evaluations per set. 

\begin{figure}[t]
\begin{center}
\centering
\includegraphics[width=1\linewidth]{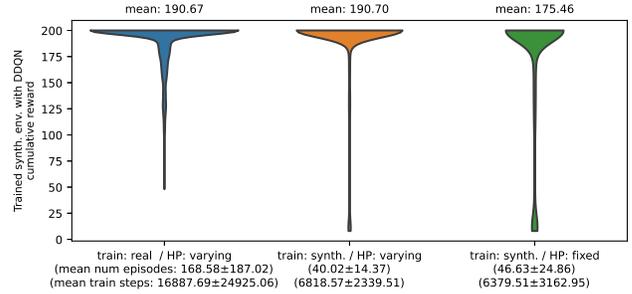} 
\end{center}
\vspace*{-0.2cm}
\caption{We show the distributions of the average cumulative rewards collected by DDQN agents on the CartPole task based on 4000 evaluations per violin and DDQN-trained SEs. We depict three different agent training settings: (left) agents trained on real environments with varying agent HPs, (center) on DDQN-trained SEs when varying agent HPs during NES runs, (right) on DDQN-trained SEs where the agent HPs were fixed during training of the SEs. The DDQN-trained SEs consistently train DDQN agents up to $\sim$60\% faster and more stably (mean train steps and std. dev. of center violin) compared to the baseline (left violin), and the agents show little sensitivity to HP variations.}
\label{fig:cartpole_ddqn_vary_hp}
\end{figure}

Lastly, we used the cumulative rewards from the test episodes for generating the violin plots seen in Figure \ref{fig:cartpole_ddqn_vary_hp}. The center violin corresponds to the set for which we trained the SEs with varying agent HPs and the right violin corresponds to the other respective set for which we did not vary the HPs. The left violin represents our baseline which shows 4000 evaluations on CartPole without an involvement of SEs. Note, in all three cases we always vary the agent HPs at test time, and the ``HP: fixed'' / ``HP: varying'' in the Figure merely indicates whether we varied the agent HPs \emph{during SE training}. We also report the average reward across the 4000 evaluations (top) and the number of average episodes and training steps required until the heuristics for stopping training (see Method Section) are triggered (bottom). We can see that the DDQN-trained SEs are consistently able to train DDQN agents using $\sim$60\% fewer steps on average while being more stably (center violin, smaller std. dev.) than the the baseline (left violin), and the SEs also show little sensitivity to HP variations. Training on SEs without varying the agent HPs during the NES optimization degrades the performances noticeably, potentially due to overfitting of the SEs to the specific agent HPs.

\subsection{Transferability of Synthetic Environments}
\begin{figure}[t]
\begin{center}
\centering
\includegraphics[width=1\linewidth]{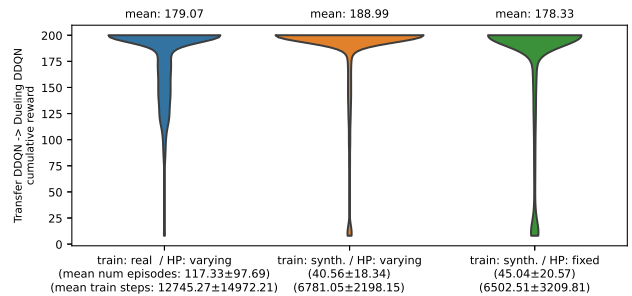} 
\end{center}
\vspace*{-0.2cm}
\caption{Visualization of transferring from DDQN to Dueling DDQN agents on CartPole based on 4000 evaluations per violin.}
\label{fig:cartpole_duelingddqn_vary_hp}
\vspace*{-0.2cm}
\end{figure}

Obviously, it is difficult to justify the argument of speed improvement when a lot of environment observations have gone into training an SE. This is why in this experiment, we investigate whether DDQN-trained SEs are capable of efficiently training other agents as well. To do this, we reuse the two sets from the previous experiment that consist of 40 DDQN-trained SEs, but this time we train Dueling DDQN agents on the SEs, again with varying HPs according to Table \ref{tab:vary_hp}. From Figure \ref{fig:cartpole_duelingddqn_vary_hp} we conclude that the transfer to the Dueling DDQN agent succeeds and it facilitates a $\sim$50\% faster and noticeably more stable training on average. Again, presumably due to overfitting, not varying the DDQN agent HPs during our NES runs negatively affects the Dueling DDQN performance (right violin).
% below the figures we show the training statistics *leading* to the test results and the violins

\begin{figure}[t]
\begin{center}
\centering
\includegraphics[width=1.\linewidth]{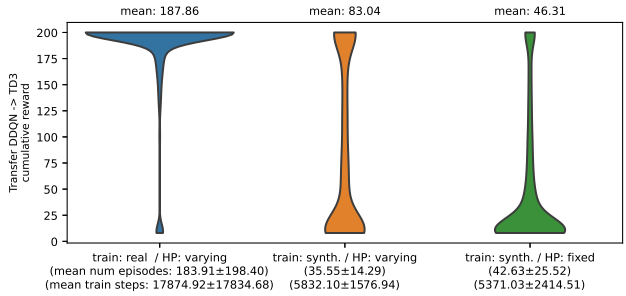} 
\end{center}
\vspace*{-0.2cm}
\caption{Visualization of transferring from DDQN to discrete-action-space TD3 on the CartPole task.}
\label{fig:cartpole_td3d_vary_hp}
\vspace*{-0.1cm}
\end{figure} 

\setcounter{figure}{4}
\begin{figure}[bt]
\begin{center}
\centering
\includegraphics[width=1\linewidth]{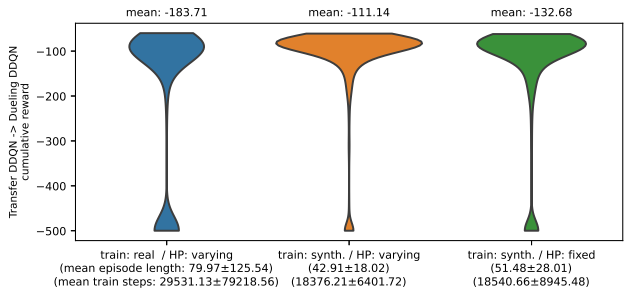} 
\end{center}
\vspace*{-0.2cm}
\caption{Visualization of transferring from DDQN to Dueling DDQN on the Acrobot task.}
\label{fig:acrobot_duelingddqn_vary_hp}
\end{figure}

\setcounter{figure}{5}
\begin{figure*}
    \centering
    \begin{minipage}{.2\textwidth}
        \centering
        \includegraphics[width=1\linewidth]{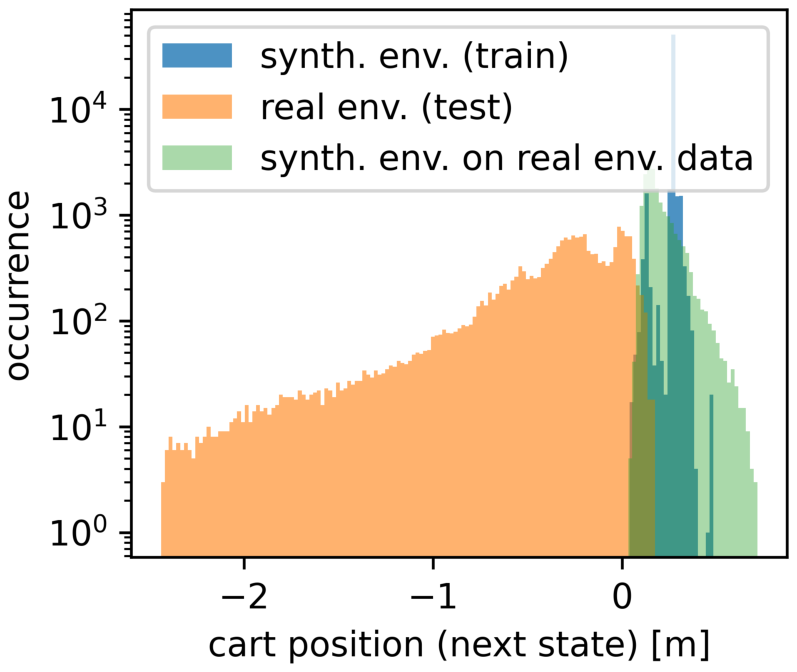}
    \end{minipage}%
    \begin{minipage}{.2\textwidth}
        \centering
        \includegraphics[width=1\linewidth]{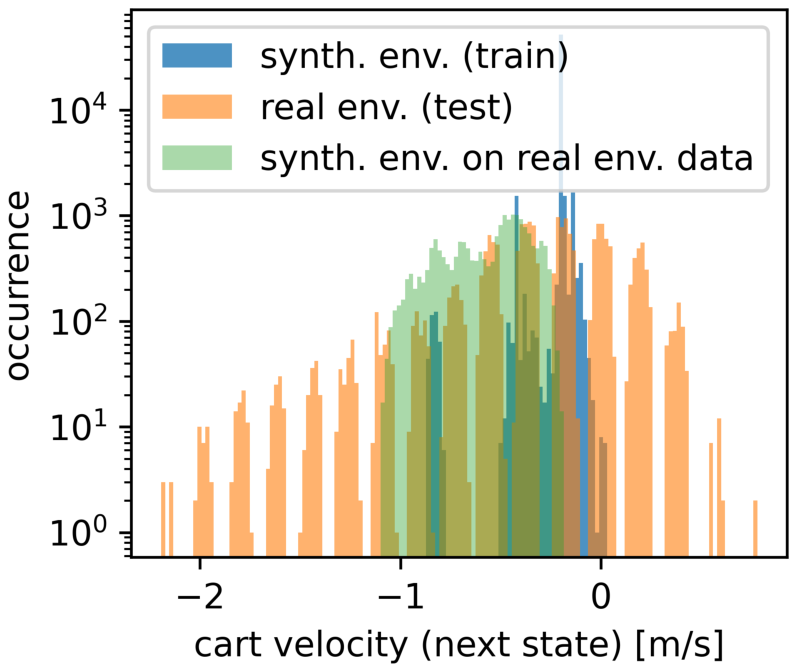}
    \end{minipage}%
    \begin{minipage}{.2\textwidth}
        \centering
        \includegraphics[width=1\linewidth]{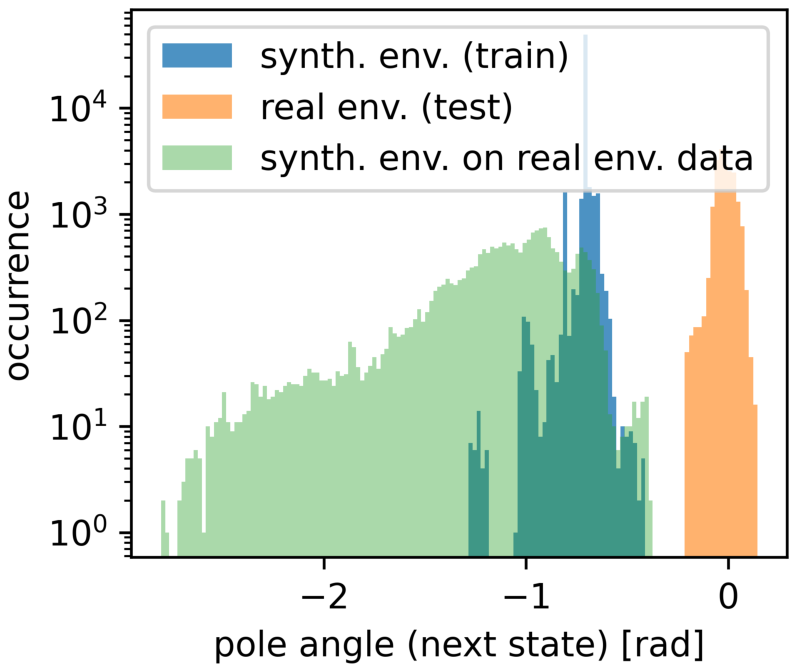}
    \end{minipage}%
    \begin{minipage}{.2\textwidth}
        \centering
        \includegraphics[width=1\linewidth]{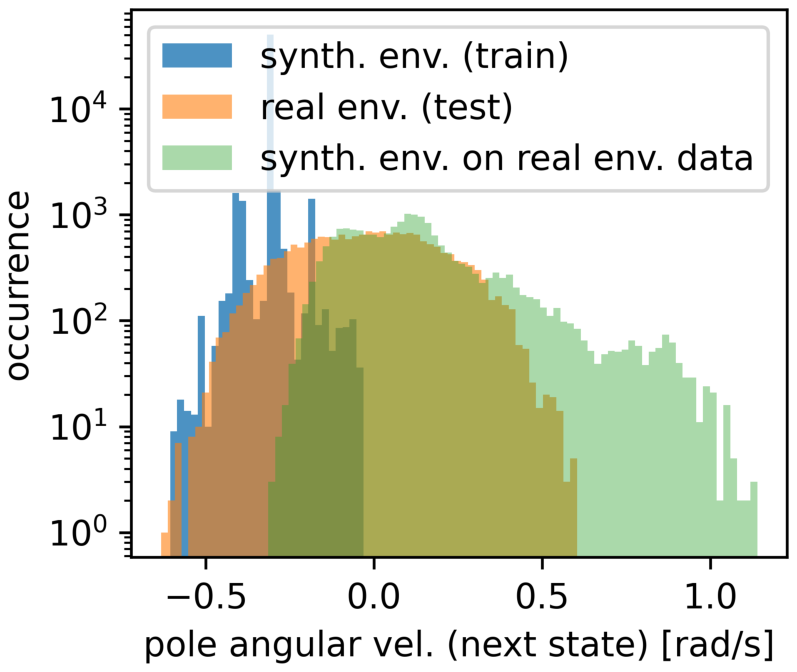}
    \end{minipage}%
    \begin{minipage}{.2\textwidth}
        \centering
        \includegraphics[width=1\linewidth]{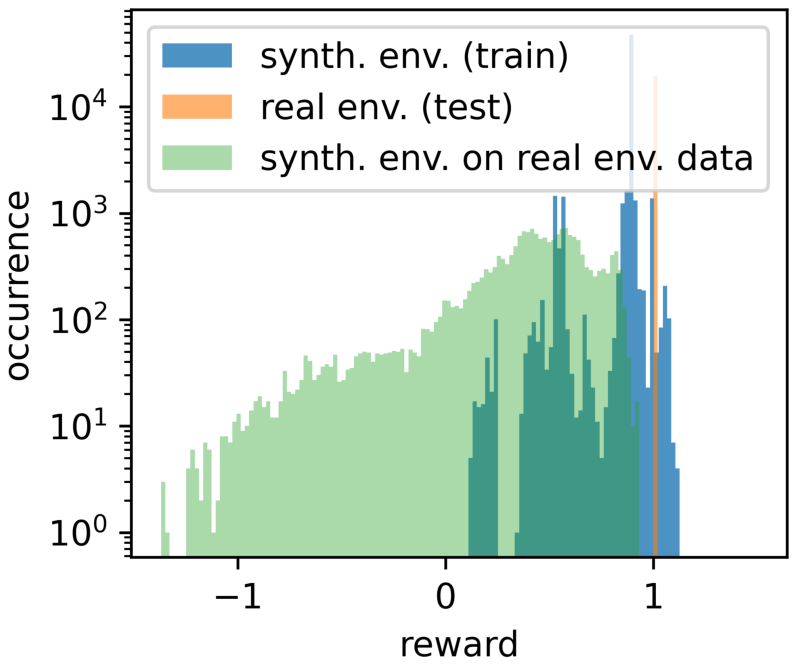}
    \end{minipage}\\
    
    % not plotting joint velocities
    \centering
    \begin{minipage}{.2\textwidth}
        \centering
        \includegraphics[width=1\linewidth]{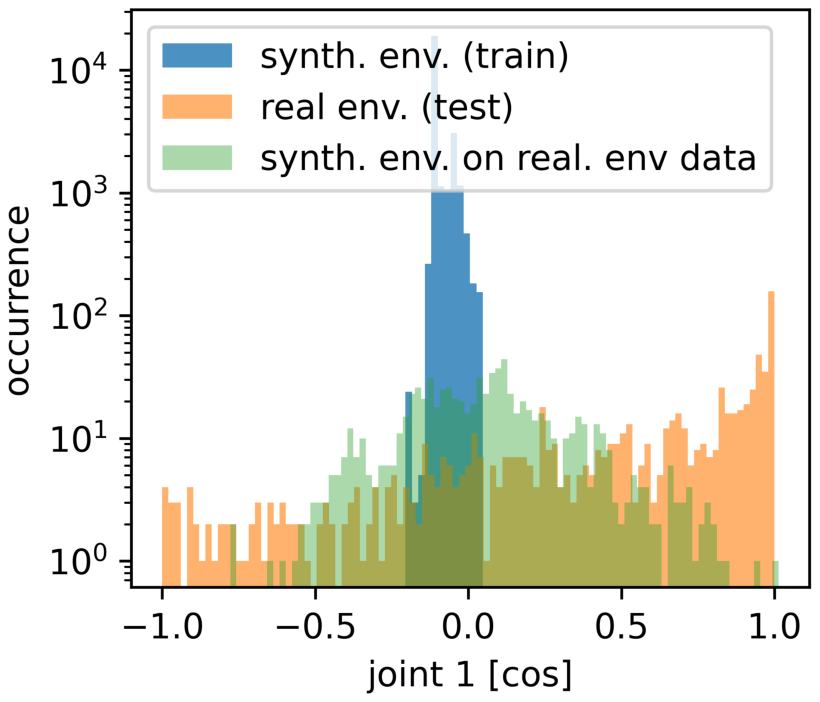}
    \end{minipage}%
    \begin{minipage}{.2\textwidth}
        \centering
        \includegraphics[width=1\linewidth]{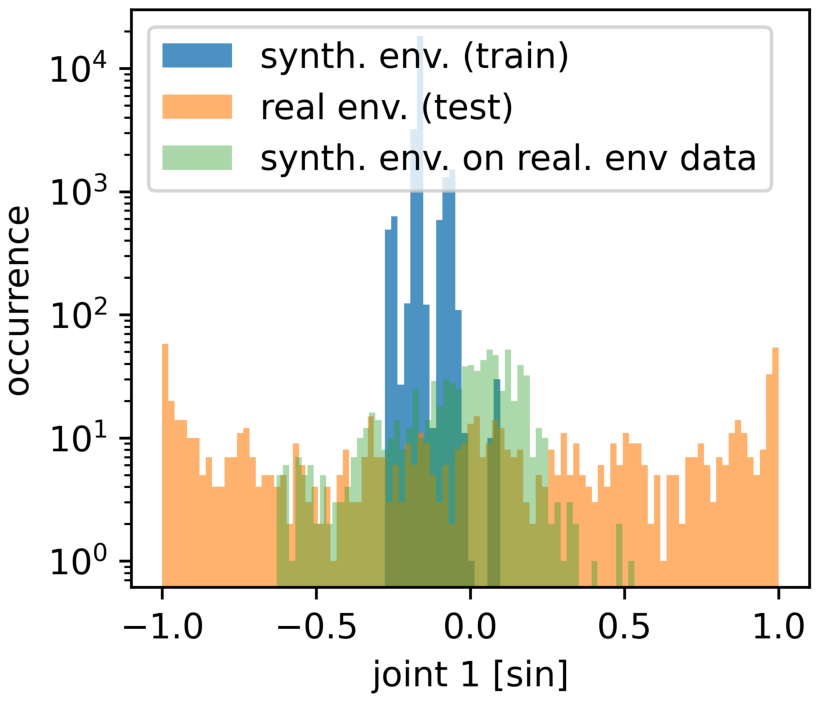}
    \end{minipage}%
    \begin{minipage}{.2\textwidth}
        \centering
        \includegraphics[width=1\linewidth]{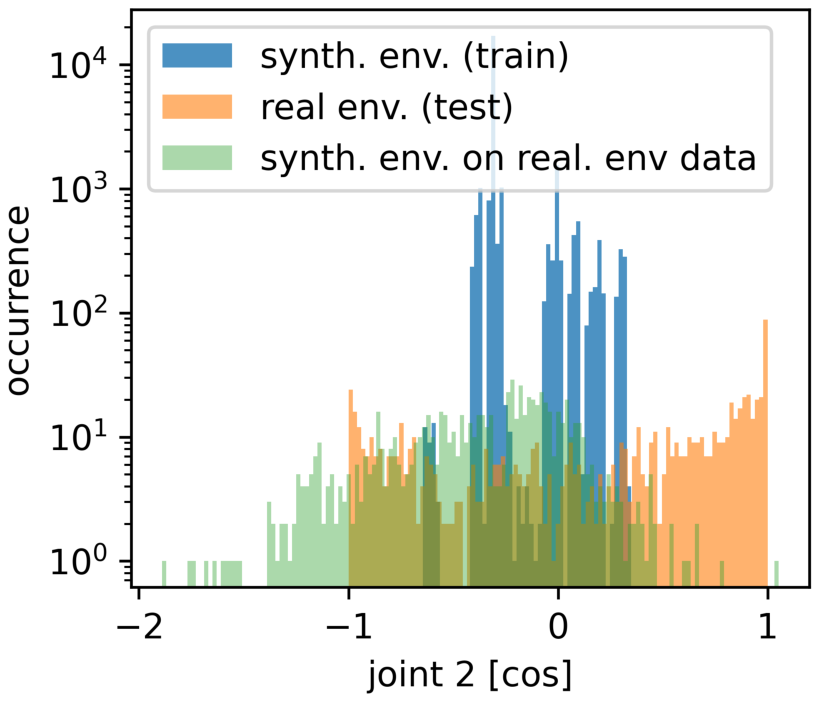}
    \end{minipage}%
    \begin{minipage}{.2\textwidth}
        \centering
        \includegraphics[width=1\linewidth]{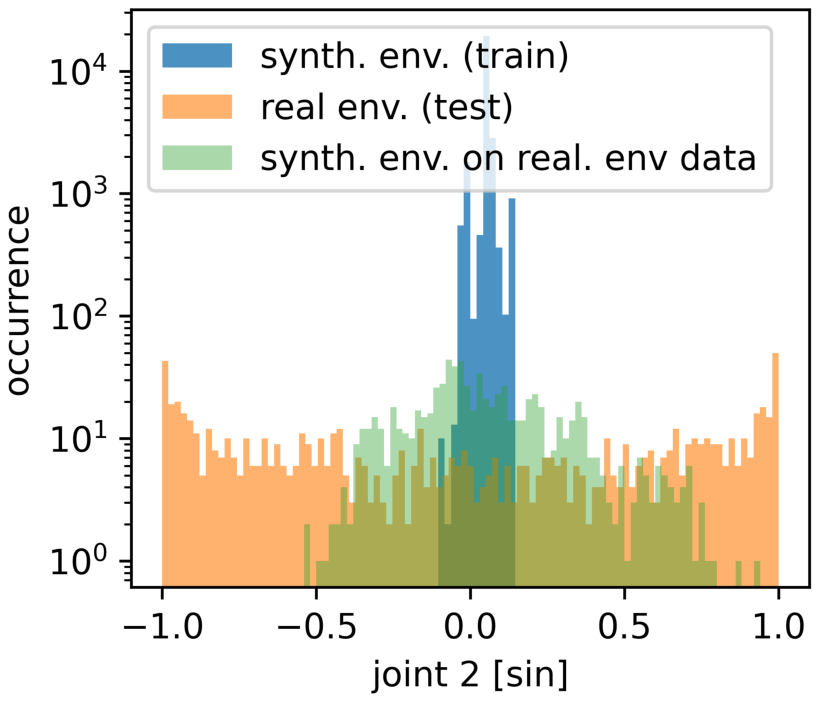}
    \end{minipage}%
    \begin{minipage}{.2\textwidth}
        \centering
        \includegraphics[width=1\linewidth]{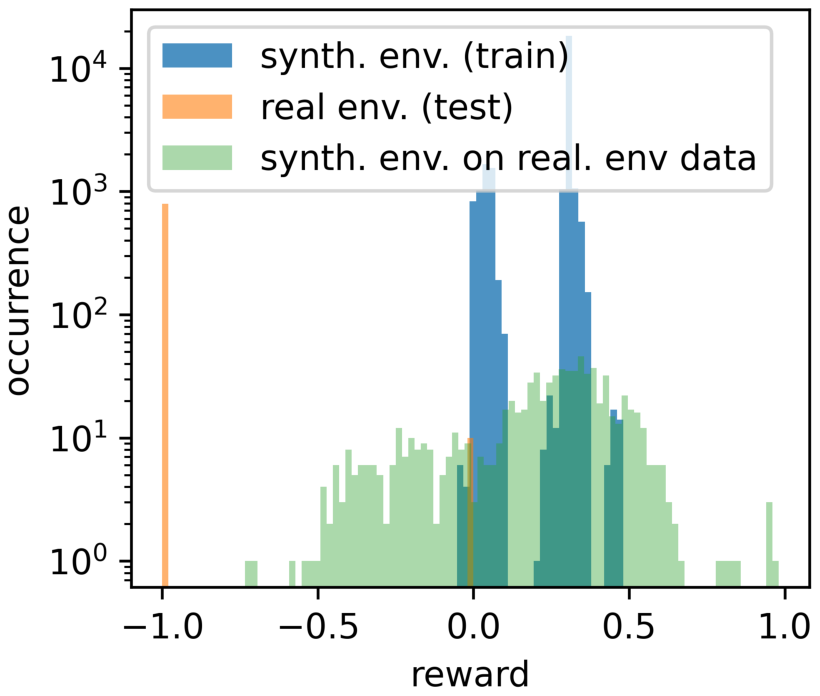}
    \end{minipage}%
    
\vspace*{-.2cm}
    \caption{Histograms of approximate next state $s'$ and reward $r$ distributions produced by 10 DDQN agents when trained on an SE (blue) and when afterwards tested for 10 episodes on a real environment (orange) for each task (top: CartPole, bottom: Acrobot). In green we depict the SE responses when the SE is fed with state-action pairs that the agent used during testing on the real environment. For CartPole, we show all state dimensions and for Acrobot we only show the joint angles.}
    \label{fig:histograms}
    
\end{figure*}

We conducted another experiment to analyze the transfer to a second agent which is not solely based on (deep) Q-learning. As described above, we chose the actor-critic-based TD3 agent that we adapted to handle discrete action spaces while maintaining differentiability. We can see in Figure \ref{fig:cartpole_td3d_vary_hp} that, while the baseline performance (left violin) indicates that our implementation seems to be correct, the performance decreases in the center violin, showing a limited transferability compared to the very good transfer to Dueling DDQN. We believe this may be due to the different learning behavior of actor-critic methods compared to learning with DDQN. We believe this result may indicate that \emph{TrainAgent} requires even more variation, i.e. instead of varying the HPs and seeds, we may additionally vary the agent types within an NES run. Another way to address this might be to increase the number of evaluations of the same perturbation and adding additional workers. Nevertheless, we also observe that in some cases the SEs are capable of training discrete TD3 agents successfully and from the low std. dev. (14.29 vs. 198.40) of the average number of episodes we can infer that, when the training succeeds, it remains efficient. All in all, this begs a deeper analysis of the transfer in the future, for example by identifying and analysing well and ill-suited SEs, as we were able to observe that well-suited SEs tended to be consistent in their aptitude of training agents.
%Lastly, it should be noted that the SE overfitting to fixed HPs again negatively affects performance (right violin). 
%
As can be seen in Figure \ref{fig:acrobot_duelingddqn_vary_hp}, transferring from DDQN to Dueling DDQN is also possible on the Acrobot task (a cumulative reward of -100 solves the task). In this case, the policies learned on the SE (center violin) are in fact substantially \emph{better} on average than those learned on the real environment (left violin). Also, the SEs facilitate a $\sim$38\% faster 
%and noticeably more stable 
training on average compared to the baseline. We note that the default DDQN HPs found for the CartPole task were reused for this task (as well as the HPs and ranges for variation from Table \ref{tab:vary_hp}). 

\subsection{Analysing Synthetic Environment Behavior}
Is it possible to shed some light on the efficacy of the learned SEs? Notice that we are operating on tasks with small state spaces which allow a qualitative visual study. This motivates the following experiment in which we visualized an approximation of the state and reward distributions generated by agents trained on SEs and real environments.

First, we randomly chose one trained SE for each task (CartPole and Acrobot). Then, on each SE we trained 10 DDQN agents with default HPs and random seeds until the stopping heuristic specified in the Method Section was triggered (similar to the other experiments). During their training, we logged all $(s, a, r, s')$ tuples. Second, we evaluated the SE-trained DDQN agents on the respective real environments for 10 test episodes each and again logged the $(s, a, r, s')$ tuples. Lastly, we visualized the histograms of the collected next states and rewards for each task and color-coded them according to their origin (SE or real environment). 

The result is depicted in Figure \ref{fig:histograms} for CartPole (top) and Acrobot (bottom). We show all four CartPole state dimensions, but only four of the six Acrobot state dimensions (only the $\sin$ and $\cos$ joint angles) for reasons of brevity. All plots show a strong distributional shift between the SE and the real environment, indicating that the agent is tested on states and provided with rewards it has barely seen during training, yet it is able to solve the environment (average cumulative rewards: 199.28 on CartPole and -90.2 on Acrobot). Furthermore, it can be observed that some of the synthetic \emph{state} distributions are narrower than the real counterparts. We point out that the synthetic \emph{reward} distribution is wider than the real one, indicating that the sparse reward distribution becomes dense as we get a reward for each action taken. We hypothesize that the SEs produce an informed representation of the target environment by narrowing the \emph{state} distributions to bias agents towards helpful (i.e. carrying strong signal) and relevant states. The histograms depicted in green were generated to see whether the distribution shifts are caused by the agent or the SE. They show the SE responses when fed with real environment data based on the logged state-action pairs that the agents have seen during testing. For most of the state dimensions, we observe that the green distributions align better with the blue than with the orange ones. Regardless of the origin (SE or real) of the current state and action, the next state and reward of the SE seem again to converge to values seen during training. Thus, we conclude it is more likely the shift is generated by the SE than the agent. While many questions remain unanswered in these preliminary results, they may offer partial explanations for the efficacy of SEs to the reader, for example, by understanding them as ``guiding systems'' for agents.

\subsection{Limitations}
Aside from advantages there also exist limitations to our approach. As can be seen in \citet{salimans-arxiv17a}, NES methods strongly depend on the number of workers and require a lot of parallel computational resources. We observed this limitation in preliminary experiments when we applied our method to more complex environments, such as the HalfCheetah or Humanoid task. Unsurprisingly, 16 workers were insufficient to learn SEs able to solve them. Moreover, we assume observable, Markovian states and partial observability may add further complexity to the optimization.

%% file: 4_conclusion.tex
\section{Conclusion}
We proposed a method that allows to learn synthetic environments which act as proxies for RL target task environments. By analyzing this method for the two discrete-action-space target tasks CartPole and Acrobot, we provided empirical evidence of their efficacy in experiments with 4000 evaluations and under varying hyperparameters and agents. Our results show that it is possible to significantly reduce the number of training steps while still achieving the same target task performance. Moreover, the results illustrate that the learned SEs are capable of transferring well (Dueling DDQN) or in limited ways (TD3) to new agents not seen during training. While SEs still have to be better understood, we see them as a useful tool in various applications. For example, as agent-agnostic, cheap-to-run environments for AutoML, as a tool for agent and task analysis or as models for efficient agent pre-training. We believe our promising results motivate future research in which we want to investigate how the method performs on more complex environments and better understand the trade-off between their complexity and the required computational resources.

% , e.g. with larger state and continuous action spaces, and on tasks requiring curriculum learning.

%and and when alternating between different agents during SE training.

%% file: 5_acknowledgements.tex
\section*{Acknowledgements} The authors acknowledge funding by Robert Bosch GmbH.

%% file: 6_appendix.tex
\appendix
\onecolumn
\section*{Appendix: Agent and NES Hyperparameters}

The following tables provide an overview of the hyperparameters used in our experiments.
\begin{table}[h!]
\footnotesize
\centering
\begin{tabular}{ |l|c|c|c|c|c| }  
 \hline
 hyperparameter & symbol & CartPole-v0 & Acrobot-v1 & value range & log. scale \\ 
 \hline
 NES step size                     & $\alpha$ & $0.148$ & $0.727$ & $0.1-1$ & True \\ 
 NES std. dev.               & $\sigma$ & $0.0124$ & $0.0114$ & $0.01-1$ & True \\ 
 NES mirrored sampling               & - & True & True & False/True & - \\ 
 NES score transformation         & - & better avg. & better avg. & (rank transform, linear transform, etc.) & - \\ 
 NES SE number of hidden layers   & - & $1$ & $1$ & $1-2$ & False \\ 
 NES SE hidden layer size   & - & $83$ & $167$ & $48-192$ & True \\ 
 NES SE activation function & - & LReLU & PReLU & Tanh/ReLU/LReLU/PRelu & - \\ 
 DDQN initial episodes                & - & $1$ & $20$ & $1-20$ & True \\ 
 DDQN batch size                      & - & $199$ & $149$ & $64-256$ & False \\ 
 DDQN learning rate                   & - & $0.000304$ & $0.00222$ & $0.0001-0.005$ & True \\ 
 DDQN target network update rate      & - & $0.00848$ & $0.0209$ & $0.005-0.05$ & True \\ 
 DDQN discount factor                 & - & $0.988$ & $0.991$ & $0.9-0.999$ & True (inv.) \\ 
 DDQN initial epsilon                 & - & $0.809$ & $0.904$ & $0.8-1$ & True \\ 
 DDQN minimal epsilon                 & - & $0.0371$ & $0.0471$ & $0.005-0.05$ & True \\ 
 DDQN epsilon decay factor            & - & $0.961$ & $0.899$ & $0.8-0.99$ & True (inv.) \\ 
 DDQN number of hidden layers         & - & $1$ & $1$ & $1-2$ & False \\ 
 DDQN hidden layer size               & - & $57$ & $112$ & $48-192$ & True \\ 
 DDQN activation function             & - & Tanh & LReLU & Tanh/ReLU/LReLU/PRelu & - \\ 
 \hline
\end{tabular} \par
\caption{Optimized hyperparameters for experiment depicted in Figure \ref{fig:cartpole_acrobot_success}}
\label{tab:exp_synth_best_performance_opt}
%\vspace*{-4cm}
\end{table}

\begin{table}[h!]
\footnotesize
\centering
\begin{tabular}{ |l|c|c|c| }  
 \hline
 hyperparameter & symbol & CartPole-v0 & Acrobot-v1 \\ 
 \hline
 NES number of outer loops         & $n_o$      & $200$     & $200$ \\
 NES max. number of train episodes & $n_e$      & $1000$   & $1000$ \\
 NES number of test episodes       & $n_{te}$   & $10$     & $10$ \\
 NES population size               & $n_p$      & $16$     & $16$ \\
 DDQN replay buffer size              & -          & $100000$ & $100000$ \\
 DDQN early out number                & $d$  & $10$     & $10$ \\
 DDQN early out difference            & $C_{diff}$ & $0.01$   & $0.01$ \\
 env. max. episode length             & -          & $200$    & $500$ \\
 env. solved reward                   & -          & $195$    & $-100$ \\
 \hline
\end{tabular} \par
\caption{Fixed hyperparameters for experiment depicted in Figure \ref{fig:cartpole_acrobot_success}}
\label{tab:exp_synth_best_performance_fixed}
%\vspace*{-4cm}
\end{table}

\begin{table}[h!]
\footnotesize
\centering
\begin{tabular}{ |l|c|c| }  
 \hline
 NES, SE, DDQN, Dueling DDQN \& TD3 hyperparameter & value \\ 
 \hline
 NES step size                       & $1$ \\
 NES std. dev.                 & $0.05$ \\
 NES mirrored sampling               & True \\ 
 NES score transformation            & better avg. \\
 NES SE number of hidden layers & $1$ \\ 
 NES SE hidden layer size   & $128$ \\ 
 NES SE activation function & LReLU \\ 
 initial episodes                             & $10$  \\ 
 batch size                       & $128$  \\ 
 learning rate (DDQN \& D.DDQN / TD3)            & $0.001$ / $0.0005$ \\ 
 target network update rate               & $0.01$ \\ 
 discount factor                          & $0.99$  \\ 
 epsilon decay factor              & $0.9$ \\ 
 number of hidden layers (agents)                  & $2$  \\ 
 hidden layer size (agents)                      & $128$ \\ 
 activation function (DDQN \& D. DDQN / TD3)          & ReLU / Tanh \\ 
 replay buffer size            & $100000$ \\
 max. train episodes            & $1000$ \\
 Gumbel Softmax start temperature / one-hot-encoded actions (TD3)        & $1$ / False \\
 \hline
\end{tabular} \par
\caption{Default HPs for experiments used in Fig. \ref{fig:cartpole_ddqn_vary_hp}, \ref{fig:cartpole_duelingddqn_vary_hp}, \ref{fig:cartpole_td3d_vary_hp}, \ref{fig:acrobot_duelingddqn_vary_hp}, and \ref{fig:histograms}. \emph{early out num} and \emph{early out diff.} are equivalent to Table \ref{tab:exp_synth_best_performance_fixed}.}
\label{tab:default_hps}
\end{table}